\documentclass[a4paper,final]{article}
\usepackage[dvips]{graphicx}
\usepackage{amsmath} 
\usepackage{amssymb} 

\makeatletter

\renewenvironment{abstract}
  {\normalfont
    \list{}{\labelwidth0pt
      \leftmargin0pt \rightmargin\leftmargin
      \listparindent\parindent \itemindent0pt
      \parsep0pt
      
    }
    \item[\hskip\labelsep\bfseries\abstractname\enspace --] \itshape
}{
  \endlist}

\newcommand{\keywordsname}{Keywords}
\newenvironment{keywords}
  {\normalfont
    \list{}{\labelwidth0pt
      \leftmargin0pt \rightmargin\leftmargin
      \listparindent\parindent \itemindent0pt
      \parsep0pt
      }
    \item[\hskip\labelsep\bfseries\keywordsname:]}{\endlist}

\begin{document}

\pagestyle{myheadings}
\markboth{}{}

\title{Qualitative Belief Conditioning Rules (QBCR)}

\author{
Florentin Smarandache\\
Department of Mathematics\\
University of New Mexico\\
Gallup, NM 87301, U.S.A.\\
smarand@unm.edu\\
\and Jean Dezert\\
ONERA\\
29 Av. de la  Division Leclerc \\
92320 Ch\^{a}tillon, France.\\
Jean.Dezert@onera.fr
}

\date{}

\maketitle

\begin{abstract}
In this paper we extend the new family of (quantitative) Belief Conditioning Rules (BCR) recently developed in the Dezert-Smarandache Theory (DSmT) to their qualitative counterpart for belief revision. Since the revision of quantitative as well as qualitative belief assignment given the occurrence of a new event (the conditioning constraint) can be done in many possible ways, we present here only what we consider as the most appealing Qualitative Belief Conditioning Rules (QBCR) which allow to revise the belief directly with words and linguistic labels and thus avoids the introduction of ad-hoc translations of quantitative beliefs into quantitative ones for solving the problem.
\end{abstract}


\section{Introduction}
\label{sec:Introduction}

In this paper, we propose a simple arithmetic of linguistic labels which allows a direct extension of quantitative Belief Conditioning Rules (BCR) proposed in the DSmT \cite{DSmTBook_2004,DSmTBook_2006} framework to their qualitative counterpart. Qualitative beliefs assignments are well adapted for manipulated information expressed in natural language and usually reported by human expert or AI-based expert systems. A new method for computing directly with words (CW) for combining and conditioning qualitative information is presented. CW, more precisely computing with linguistic labels, is usually more vague, less precise than computing with numbers, but it is expected to offer a better robustness and flexibility for combining uncertain and conflicting human reports than computing with numbers because in most of cases human experts are less efficient to provide (and to justify) precise quantitative beliefs than qualitative beliefs. 

Before extending the quantitative DSmT-based conditioning rules to their qualitative counterparts, it will be necessary to define few but new important operators on linguistic labels and what is a qualitative belief assignment. Then we will show though simple examples how the combination of qualitative beliefs can be obtained in the DSmT framework.

\section{Qualitative operators and belief assignments}
\label{sec2}
Since one wants to compute directly with words (CW) instead of numbers, we define without loss of generality a finite set of linguistic labels $\tilde{L}=\{L_1,L_2,\ldots,L_n\}$ where $n\geq 2$ is an integer. $\tilde{L}$ is endowed with a total order relationship $\prec$, so that $L_1\prec L_2\prec \ldots\prec L_n$. To work on a close linguistic set under linguistic addition and multiplication operators, one extends $\tilde{L}$ with two extreme values $L_{0}$ and $L_{n+1}$ where $L_{0}$ corresponds to the minimal qualitative value and $L_{n+1}$ corresponds to the maximal qualitative value, in such a way that $L_0\prec L_1\prec L_2\prec \ldots\prec L_n\prec L_{n+1}$ where $\prec$ means inferior to, or less, or smaller (in quality) than, etc. Therefore, one will work on the extended ordered set $L$ of qualitative values $L=\{L_0,L_1,L_2,\ldots,L_n,L_{n+1}\}$. The qualitative addition and multiplication of linguistic labels, which are commutative, associative, and unitary operators,  are defined as follows - see Chapter 10 in \cite{DSmTBook_2006} for details and examples :
\begin{itemize}
\item Addition : if  $i+j < n+1$, $L_i + L_j=L_{i+j}$ otherwise $L_i + L_j=L_{n+1}$.
\item Multiplication\footnote{A more precise multiplication operator has been proposed in \cite{Li2007}.} : $L_i \times L_j=L_{\min\{i,j\}}$
\end{itemize}

Let's consider a finite and discrete frame of discernment $\Theta=\{\theta_1,\ldots,\theta_n\}$ for the given problem under consideration where the true solution must lie in; its model $\mathcal{M}(\Theta)$ defined by the set of integrity constraints on elements of $\Theta$ (i.e. free-DSm model, hybrid model or Shafer's model) and its corresponding hyper-power set denoted $D^\Theta$; that is, the Dedekind's lattice on $\Theta$ \cite{DSmTBook_2004} which is nothing but the space of propositions generated with $\cap$ and $\cup$ operators and elements of $\Theta$ taking into account the integrity constraints (if any) of the model.  A qualitative basic belief assignment (qbba) also called qualitative belief mass is a mapping function $qm(.): D^\Theta \mapsto L$. In the sequel, all qualitative masses not explicitly specified in the examples, are by default (and for notation convenience) assumed to take the minimal linguistic value $L_0$.

\section{Quasi-normalization of qualitative masses}

There is no way to define a normalized $qm(.)$, but a qualitative quasi-normalization \cite{DSmTBook_2006} is nevertheless possible if needed as follows:
\begin{itemize}
\item[a)] If the previous defined labels $L_0$, $L_1$, $L_2$, $\ldots$, $L_n$, $L_{n+1}$ from the set $L$ are equidistant, i.e. the (linguistic) distance between any two consecutive labels $L_j$ and $L_{j+1}$ is the same, for any $j \in \{0, 1, 2, \ldots, n\}$, then one can make an isomorphism between $L$ and a set of sub-unitary numbers from the interval $[0, 1]$ in the following way:  $L_i = i/(n+1)$, for all $i \in \{0, 1, 2, \ldots, n+1\}$, and therefore the interval $[0, 1]$ is divided into $n+1$ equal parts. Hence, a qualitative mass, $qm(X_i) = L_i$, is equivalent to a quantitative mass $m(X_i) = i/(n+1)$ which is normalized if 
$$\sum_{X\in D^\Theta} m(X)= \sum_{k} i_k/(n+1)=1$$
\noindent
but this one is equivalent to 
$$\sum_{X\in D^\Theta} qm(X)= \sum_{k} L_{i_k}=L_{n+1}$$
\noindent In this case we have a {\it{qualitative normalization}}, similar to the (classical) numerical normalization.
\item[b)] But, if the previous defined labels $L_0$, $L_1$, $L_2$, $\ldots$, $L_n$, $L_{n+1}$ from the set $L$ are not equidistant, so the interval $[0, 1]$ cannot be split into equal parts according to the distribution of the labels, then it makes sense to consider a {\it{qualitative quasi-normalization}}, i.e. an approximation of the (classical) numerical normalization for the qualitative masses in the same way:
 $$\sum_{X\in D^\Theta} qm(X)=L_{n+1}$$
\noindent
In general, if we don't know if the labels are equidistant or not, we say that a qualitative mass is quasi-normalized when the above summation holds.
\end{itemize}

\section{Quantitative Belief Conditioning Rules (BCR)}

Before presenting the new Qualitative Belief Conditioning Rules (QBCR) in the next section, it is first important and necessary to briefly recall herein what are the (quantitative) Belief Conditioning Rules (BCR) and what was the motivation for their development in DSmT framework and also the fundamental difference between BCR and Shafer's Conditioning Rule (SCR) proposed in \cite{Shafer_1976}.\\

So, let's suppose one has a prior basic belief assignment (bba) $m(.)$ defined on hyper-power set $D^\Theta$, and one finds out (or one assumes) that the truth is in a given element $A\in D^\Theta$, i.e. $A$ has really occurred or is supposed to have occurred. The problem of belief conditioning is on how to revise properly the prior bba $m(.)$ with the knowledge about the occurrence of $A$. Simply stated: how to compute $m(.|A)$ from the knowledge available, that is with any prior bba $m(.)$ and $A$ ?

\subsection{Shafer's Conditioning Rule (SCR)}

Until very recently, the most commonly used conditioning rule for belief revision was the one proposed by Shafer \cite{Shafer_1976} and referred here as Shafer's Conditioning Rule (SCR). The SCR consists in combining the prior bba $m(.)$ with a specific bba focused on $A$ with Dempster's rule of combination for transferring the conflicting mass to non-empty sets in order to provide the revised bba. In other words, the conditioning by a proposition $A$, is obtained by SCR as follows :

\begin{equation}
m_{SCR}(.|A)=[m\oplus m_S] (.)
\label{eqSCR}
\end{equation}

\noindent
where $m(.)$ is the prior bba to update, $A$ is the conditioning event, $m_S(.)$ is the bba focused on $A$ defined by $m_S(A)=1$ and $m_S(X)=0$ for all $X\neq A$ and $\oplus$ denotes the Dempster's rule of combination \cite{Shafer_1976}.\\

The SCR approach based on Dempster's rule of combination of the prior bba with the bba focused on the conditioning event remains {\it{subjective}} since actually in such belief revision process both sources are subjective and SCR doesn't manage properly the objective nature/absolute truth carried by the conditioning term. Indeed, when conditioning a prior mass $m(.)$, {\it{knowing}} (or assuming) that the truth is in $A$, means that we have in hands an absolute (not subjective) knowledge, i.e. the truth in $A$ has occurred (or is assumed to have occurred), thus $A$ is realized (or is assumed to be realized) and this is (or at least must be interpreted as) an absolute truth. The conditioning term "Given $A$" must therefore be considered as an absolute truth, while $m_S(A)=1$ introduced in SCR cannot refer to an absolute truth actually, but only to a {\it{subjective certainty}} on the possible occurrence of $A$ from a {\it{virtual}} second source of evidence.  The advantage of SCR remains undoubtedly in its simplicity and the main argument in its favor is its coherence with conditional probability when manipulating Bayesian belief assignment. But in our opinion, SCR should better be interpreted as the fusion of $m(.)$ with a particular subjective bba $m_S(A)=1$ rather than an objective belief conditioning rule. This fundamental remark motivated us to develop a new family of BCR \cite{DSmTBook_2006} based on hyper-power set decomposition (HPSD) explained briefly in the next section. It turns out that many BCR are possible because the redistribution of masses of elements outside of $A$ (the conditioning event) to those inside $A$ can be done in $n$-ways.  This will be briefly presented right after the next section.

\subsection{Hyper-Power Set Decomposition (HPSD)}

Let $\Theta=\{\theta_1,\theta_2,\ldots,\theta_n\}$, $n\geq 2$, a model $\mathcal{M}(\Theta)$ associated for $\Theta$ (free DSm model, hybrid or Shafer's model) and its corresponding hyper-power set $D^\Theta$. Let's consider a (quantitative) basic belief assignment (bba) $m(.): D^\Theta \mapsto [0,1]$ such that $\sum_{X\in D^\Theta}m(X)=1$. Suppose one finds out that the truth is in the set $A\in D^\Theta\setminus\{\emptyset\}$. Let $\mathcal{P}_{\mathcal{D}}(A)=2^A \cap D^{\Theta} \setminus \{\emptyset\}$, i.e. all non-empty parts (subsets) of $A$ which are included in $D^\Theta$. Let's consider the normal  cases when $A\neq\emptyset$ and $\sum_{Y\in \mathcal{P}_{\mathcal{D}}(A)}m(Y)> 0$. For the degenerate case when the truth is in $A=\emptyset$, we consider Smets' open-world, which means that there are other hypotheses $\Theta'=\{\theta_{n+1},\theta_{n+2},\ldots\theta_{n+m}\}$, $m\geq 1$, and the truth is in $A\in D^{\Theta'}\setminus\{\emptyset\}$. If $A=\emptyset$ and we consider a close-world, then it means that the problem is impossible. For another degenerate case, when $\sum_{Y\in \mathcal{P}_{\mathcal{D}}(A)}m(Y)=0$, i.e. when the source gave us a totally (100\%) wrong information $m(.)$, then, we define: $m(A|A)\triangleq 1$ and, as a consequence, $m(X|A)=0$ for any $X\neq A$. Let $s(A)=\{\theta_{i_1},\theta_{i_2},\ldots,\theta_{i_p}\}$, $1\leq p\leq n$,  be the singletons/atoms that compose $A$ (for example, if $A=\theta_1\cup(\theta_3\cap\theta_4)$ then $s(A)=\{\theta_1,\theta_3,\theta_4\}$). The Hyper-Power Set Decomposition (HPSD) of $D^\Theta \setminus \emptyset$ consists in its decomposition into the three following subsets generated by $A$:
\begin{itemize}
\item
$D_1=\mathcal{P}_{\mathcal{D}}(A)$, the parts of $A$ which are included in the hyper-power set, except the empty set; 
\item
$D_2=\{(\Theta\setminus s(A)),\cup , \cap\} \setminus \{\emptyset\}$, i.e. the sub-hyper-power set generated by $\Theta\setminus s(A)$ under $\cup$ and $\cap$, without the empty set.
\item
$D_3=(D^\Theta\setminus\{\emptyset\}) \setminus (D_1\cup D_2)$; each set from $D_3$ has in its formula singletons from both $s(A)$ and $\Theta\setminus s(A)$ in the case when $\Theta\setminus s(A)$ 
is different from empty set.
\end{itemize}
\noindent
$D_1$, $D_2$ and $D_3$ have no element in common two by two and their union is $D^\Theta\setminus\{\emptyset\}$.\\

\noindent
{\it{Simple example of HPSD}}: Let's consider $\Theta=\{\theta_1, \theta_2, \theta_3\}$ with Shafer's model (i.e. all elements of $\Theta$ are exclusive) and let's assume that the truth is in $\theta_2\cup \theta_3$, i.e. the conditioning term is $\theta_2\cup \theta_3$. Then one has the following HPSD: $D_1=\{\theta_2,\theta_3,\theta_2\cup \theta_3\}$, $D_2=\{\theta_1\}$ and $D_3=\{\theta_1\cup \theta_2, \theta_1\cup \theta_3, \theta_1\cup \theta_2\cup \theta_3\}$. More complex and detailed examples can be found in \cite{DSmTBook_2004}.

\subsection{Belief conditioning rules (BCR)}

Since there exists actually many ways for redistributing the masses of elements outside of $A$ (the conditioning event) to those inside $A$, several BCR have been proposed recently in \cite{DSmTBook_2006}. Due to space limitation, we will not browse here all the possibilities for doing these redistributions and all BCR but one just presents here a typical and interesting BCR, i.e. the BCR number 17 (i.e. BCR17) which does in our opinion the most refined redistribution since:
\newline - the mass $m(W)$ of each element $W$ in $D_2\cup D_3$ is transferred to those $X\in D_1$ elements which are
included in $W$ if any proportionally with respect to their non-empty masses;
\newline - if no such $X$ exists, the mass $m(W)$ is transferred in a pessimistic/prudent way to the $k$-largest element from $D_1$ which are included in $W$ (in equal parts) if any;
\newline - if neither this way is possible, then $m(W)$ is indiscriminately distributed to all $X \in D_1$
proportionally with respect to their nonzero masses.\\

BCR17 is defined by the following formula (see \cite{DSmTBook_2004}, Chap. 9 for detailed explanations and examples):

\begin{equation}
m_{BCR17}(X|A)=
m(X)\cdot \Bigg[ S_{D_1}
+ 
\displaystyle\sum_{
\begin{array}{c}
\scriptstyle W\in D_2 \cup D_3\\
\scriptstyle X\subset W\\
\scriptstyle S(W)\neq 0
\end{array}
} 
\frac{m(W)}{S(W)}
\Bigg]\\
+  \displaystyle\sum_{
\begin{array}{c}
\scriptstyle W\in D_2\cup D_3\\
\scriptstyle X\subset W, \, $X$ \,\text{is $k$-largest}\\
\scriptstyle S(W)=0
\end{array}}
m(W)/k
\label{eq:BCR17}
\end{equation}

\noindent
where "$X\,\text{is $k$-largest}$" means that $X$ is the $k$-largest (with respect to inclusion) set included in $W$ and

$$S(W) \triangleq \sum_{Y\in D_1,Y\subset W} m(Y)
\qquad\text{and}\qquad
S_{D_1} \triangleq \frac{1}{\sum_{Y\in D_1}m(Y)} \times
\displaystyle
\sum_{
\begin{array}{c}
\scriptstyle Z\in D_1,\\
\scriptstyle \text{or}\, Z\in D_2 \,\mid\, \nexists Y\in D_1\, \text{with}\, Y\subset Z
\end{array}} 
m(Z)$$

\noindent
{\it{A simple example for BCR17}}: Let's consider $\Theta=\{\theta_1, \theta_2, \theta_3\}$ with Shafer's model (i.e. all elements of $\Theta$ are exclusive) and let's assume that the truth is in $\theta_2\cup \theta_3$, i.e. the conditioning term is $A\triangleq \theta_2\cup \theta_3$. Then one has the following HPSD: $$D_1=\{\theta_2,\theta_3,\theta_2\cup \theta_3\}, \qquad D_2=\{\theta_1\}$$ 
$$D_3=\{\theta_1\cup \theta_2, \theta_1\cup \theta_3, \theta_1\cup \theta_2\cup \theta_3\}.$$

\noindent
Let's consider the following prior bba:
$m(\theta_1)=0.2$, $m(\theta_2)=0.1$, $m(\theta_3)=0.2$, $m(\theta_1\cup \theta_2)=0.1$, $m(\theta_2\cup \theta_3)=0.1$ and $m(\theta_1\cup \theta_2\cup \theta_3)=0.3$.\\

With BCR17, for $D_2$, $m(\theta_1)=0.2$ is transferred proportionally to all elements of $D_1$, i.e. $\frac{x_{\theta_2}}{0.1}=\frac{y_{\theta_3}}{0.2}=\frac{z_{\theta_2\cup \theta_3}}{0.1}=\frac{0.2}{0.4}=0.5$ whence the parts of $m(\theta_1)$ redistributed to $\theta_2$, $\theta_3$ and $\theta_2\cup\theta_3$ are respectively $x_{\theta_2}=0.05$, $y_{\theta_3}=0.10$, and $z_{\theta_2\cup \theta_3}=0.05$. For $D_3$, there is actually no need to transfer $m(\theta_1\cup \theta_3)$ because $m(\theta_1\cup \theta_3)=0$ in this example; whereas $m(\theta_1\cup \theta_2)=0.1$ is transferred to $\theta_2$ (no case of $k$-elements herein); $m(\theta_1\cup \theta_2\cup \theta_3)=0.3$ is transferred to $\theta_2$, $\theta_3$ and $\theta_2\cup \theta_3$ proportionally to their corresponding masses: $$\frac{x_{\theta_2}}{0.1}=\frac{y_{\theta_3}}{0.2}=\frac{z_{\theta_2\cup \theta_3}}{0.1}=\frac{0.3}{0.4}=0.75$$
\noindent
whence $x_{\theta_2}=0.075$, $y_{\theta_3}=0.15$, and $z_{\theta_2\cup \theta_3}=0.075$. Finally, one gets 
\begin{align*}
& m_{BCR17}(\theta_2|\theta_2\cup \theta_3)=0.10+0.05+0.10+0.075=0.325\\
& m_{BCR17}(\theta_3|\theta_2\cup \theta_3)=0.20+0.10+0.15=0.450\\
& m_{BCR17}(\theta_2\cup \theta_3|\theta_2\cup \theta_3)=0.10+0.05+0.075=0.225
\end{align*}

\noindent
which is different from the result obtained with SCR, since one gets in this example:

\begin{align*}
& m_{SCR}(\theta_2|\theta_2\cup \theta_3)=0.25\\
& m_{SCR}(\theta_3|\theta_2\cup \theta_3)=0.25\\
& m_{SCR}(\theta_2\cup \theta_3|\theta_2\cup \theta_3)=0.50
\end{align*}

\noindent
More complex and detailed examples can be found in \cite{DSmTBook_2004}.

\section{Qualitative belief conditioning rules (QBCR)}

In this section we  propose two Qualitative belief conditioning rules (QBCR) which extend the principles of quantitative BCR in the qualitative domain using the operators on linguistic labels defined in section \ref{sec2}. We consider from now on a general frame $\Theta=\{\theta_1,\theta_2,\ldots,\theta_n\}$, a given model $\mathcal{M}(\Theta)$ with its hyper-power set $D^\Theta$ and a given extended ordered set $L$ of qualitative values $L=\{L_0,L_1,L_2,\ldots,L_m,L_{m+1}\}$. The prior qualitative basic belief assignment (qbba) taking its values in $L$ is denoted $qm(.)$. We assume in the sequel that the conditioning event is $A\neq\emptyset$, $A\in D^\Theta$, i.e. the absolute truth is in $A$.

\subsection{Qualitative Belief Conditioning Rule no 1 (QBCR1)}

The first QBCR, denoted QBCR1, does the redistribution of masses in a
pessimistic/prudent way, as follows:
\begin{itemize}
\item transfer the mass of each element $Y$ in $D_2\cup D_3$ to the
largest element $X$ in $D_1$ which is contained by $Y$;
\item if no such $X$ element exists, then the mass of $Y$ is
transferred to $A$.
\end{itemize}

The mathematical formula for QBCR1 is then given by:
\begin{itemize}
\item If $X\notin D_1$, 
\begin{equation}
qm_{QBCR1}(X|A)=L_{\min}\equiv L_0
\end{equation}
\item If $X \in D_1$, 
\begin{equation}
qm_{QBCR1}(X|A)=qm(X) + qS_1(X,A) + qS_2(X,A)
\label{eqQBCR1}
\end{equation}
\end{itemize}

\noindent
where the addition operator involved in \eqref{eqQBCR1} corresponds to the {\it{addition operator on linguistic labels}} defined in section \ref{sec2} and where the {\it{qualitative summations}} $qS_1(X,A)$ and $qS_2(X,A)$ are defined by:

\begin{equation}
qS_1(X,A)\triangleq \displaystyle\sum_{
\begin{array}{c}
\scriptstyle Y\in D_2 \cup D_3\\
\scriptstyle X\subset Y\\
\scriptstyle X=\max
\end{array}
} 
qm(Y)
\label{eq:qS1}
\end{equation}

\begin{equation}
qS_2(X,A)\triangleq \displaystyle\sum_{
\begin{array}{c}
\scriptstyle Y\in D_2 \cup D_3\\
\scriptstyle Y\cap A=\emptyset\\
\scriptstyle X=A
\end{array}
} 
qm(Y)
\label{eq:qS2}
\end{equation}

\noindent
$qS_1(X,A)$ corresponds to the transfer of qualitative mass of each element $Y$ in $D_2\cup D_3$ to the largest element $X$ in $D_1$ and $qS_2(X,A)$ corresponds to the transfer of the mass of $Y$ is to $A$ when no such largest element $X$ in $D_1$ exists.

\subsection{Qualitative Belief Conditioning Rule no 2 (QBCR2)}

The second QBCR, denoted QBCR2, does a uniform redistribution of masses, as follows:
\begin{itemize}
\item transfer the mass of each element $Y$ in $D_2\cup D_3$ to the
largest element $X$ in $D_1$ which is contained by $Y$ (as QBCR1 does);
\item if no such $X$ element exists, then the mass of $Y$ is uniformly redistributed to all subsets of $A$ whose (qualitative) masses are not $L_0$ (i.e. to all qualitative focal elements included in $A$).
\item if there is no qualitative focal element included in $A$, then the mass of $Y$ is transferred to $A$.
\end{itemize}

The mathematical formula for QBCR2 is then given by:
\begin{itemize}
\item If $X\notin D_1$, 
\begin{equation}
qm_{QBCR2}(X|A)=L_{\min}\equiv L_0
\end{equation}
\item If $X \in D_1$, 
\begin{equation}
qm_{QBCR2}(X|A)=qm(X) + qS_1(X,A) + qS_3(X,A)+ qS_4(X,A)
\label{eqQBCR2}
\end{equation}
\end{itemize}

\noindent
where the addition operator involved in \eqref{eqQBCR2} corresponds to the {\it{addition operator on linguistic labels}} defined in section \ref{sec2} and where the {\it{qualitative summations}} $qS_1(X,A)$ is defined in \eqref{eq:qS1}, $qS_3(X,A)$ and $qS_4(X,A)$ by:

\begin{equation}
qS_3(X,A)\triangleq \displaystyle\sum_{
\begin{array}{c}
\scriptstyle Y\in D_2 \cup D_3\\
\scriptstyle Y\cap A=\emptyset\\
\scriptstyle q_F\neq 0
\end{array}
} 
\frac{qm(Y)}{q_F}
\label{eq:qS3}
\end{equation}

\begin{equation}
qS_4(X,A)
\triangleq \displaystyle\sum_{
\begin{array}{c}
\scriptstyle Y\in D_2 \cup D_3\\
\scriptstyle Y\cap A=\emptyset\\
\scriptstyle X=A, q_F=0
\end{array}
} 
qm(Y),
\label{eq:qS4}
\end{equation}

\noindent where $q_F\triangleq \text{Card}\{Z| Z\subset A, qm(Z)\neq L_0\}=$ number of qualitative focal elements of $A$.\\

\subsubsection*{Scalar division of linguistic label}

For the complete derivation of \eqref{eqQBCR2} we need to define the scalar division of labels involved in \eqref{eq:qS3}. We propose the following definition:

\begin{equation}
\frac{L_i}{j}\triangleq L_{[\frac{i}{j}]}
\label{eq:qdivision}
\end{equation}

\noindent
for all $i\geq 0$ and $j>0$ where $[\frac{i}{j}]$ is the integer part of $\frac{i}{j}$, i.e. the largest integer less than or equal to $\frac{i}{j}$. For example, $\frac{L_5}{3}= L_{[\frac{5}{3}]}=L_1$, or $\frac{L_6}{3}= L_{[\frac{6}{3}]}=L_2$, etc.

\section{Examples for QBCR1 and QBCR2}

Let's consider the following set of ordered linguistic labels\index{linguistic labels} $L=\{L_0,L_1,L_2,L_3,L_4,L_5,L_6\}$ (for example, $L_1$, $L_2$, $L_3$, $L_4$ and $L_5$ may represent the values: $L_1\triangleq \text{{\it{very poor}}}$, $L_2\triangleq \text{{\it{poor}}}$, $L_3\triangleq \text{{\it{medium}}}$, $L_4\triangleq \text{{\it{good}}}$ and $L_5\triangleq \text{{\it{very good}}}$, where  the symbol $\triangleq$ means  {\it{by definition}}). The addition and multiplication tables corresponds respectively to Tables \ref{CWTable3} and \ref{CWTable4}.
\begin{table}[h,t]
\centering 
\begin{tabular}{|c|ccccccc|}
\hline
$+$     & $L_0$ & $L_1$ & $L_2$ & $L_3$ & $L_4$ & $L_5$ & $L_6$\\
\hline
$L_0$ & $L_0$ & $L_1$ & $L_2$ & $L_3$ & $L_4$ & $L_5$ & $L_6$\\
$L_1$ & $L_1$ & $L_2$ & $L_3$ & $L_4$ & $L_5$ & $L_6$& $L_6$\\
$L_2$ & $L_2$ & $L_3$ & $L_4$ & $L_5$ & $L_6$ & $L_6$& $L_6$\\
$L_3$ & $L_3$ & $L_4$ & $L_5$ & $L_6$ & $L_6$ & $L_6$& $L_6$\\
$L_4$ & $L_4$ & $L_5$ & $L_6$ & $L_6$ & $L_6$ & $L_6$& $L_6$\\
$L_5$ & $L_5$ & $L_6$ & $L_6$ & $L_6$ & $L_6$ & $L_6$& $L_6$\\
$L_6$ & $L_6$ & $L_6$ & $L_6$ & $L_6$ & $L_6$ & $L_6$& $L_6$\\
\hline
\end{tabular}
\caption{Addition table}
\label{CWTable3}
\end{table}
\begin{table}[h,t]
\centering 
\begin{tabular}{|c|ccccccc|}
\hline
$\times$ & $L_0$ & $L_1$ & $L_2$ & $L_3$ & $L_4$ & $L_5$ & $L_6$\\
\hline
$L_0$ & $L_0$ & $L_0$ & $L_0$ & $L_0$ & $L_0$ & $L_0$& $L_0$\\
$L_1$ & $L_0$ & $L_1$ & $L_1$ & $L_1$ & $L_1$ & $L_1$& $L_1$\\
$L_2$ & $L_0$ & $L_1$ & $L_2$ & $L_2$ & $L_2$ & $L_2$& $L_2$\\
$L_3$ & $L_0$ & $L_1$ & $L_2$ & $L_3$ & $L_3$ & $L_3$& $L_3$\\
$L_4$ & $L_0$ & $L_1$ & $L_2$ & $L_3$ & $L_4$ & $L_4$& $L_4$\\
$L_5$ & $L_0$ & $L_1$ & $L_2$ & $L_3$ & $L_4$ & $L_5$& $L_5$\\
$L_6$ & $L_0$ & $L_1$ & $L_2$ & $L_3$ & $L_4$ & $L_5$& $L_6$\\
\hline
\end{tabular}
\caption{Multiplication table}
\label{CWTable4}
\end{table}

\subsection{Example 1}

\begin{figure}[h]
\centering
{\tt \setlength{\unitlength}{1pt}
\begin{picture}(90,90)
\thinlines    
\put(20,60){\circle{40}} 
\put(40,60){\circle{40}} 
\put(40,10){\circle{40}} 
\put(60,35){\circle{40}} 
\put(-5,84){\vector(1,-1){10}}
\put(-13,84){$A$}
\put(64,84){\vector(-1,-1){10}}
\put(65,84){$B$}
\put(75,10){\vector(-1,0){15}}
\put(77,7){$C$}
\put(95,35){\vector(-1,0){15}}
\put(97,34){$D$}
\end{picture}}
\vspace{2mm}
\caption{Venn Diagram for the hybrid model of Example 1 }
\label{venn1}
 \end{figure}
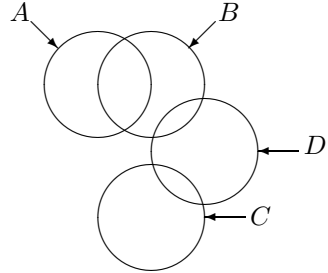

Let's consider the frame $\Theta=\{A,B,C,D\}$ with the hybrid model corresponding to the Venn diagram on Figure \ref{venn1}. We assume that the prior qualitative bba $qm(.)$ is given by:

$$qm(A)=L_1, \quad qm(C)=L_1, \quad qm(D)=L_4$$

\noindent
and the qualitative masses of all other elements of $G^\Theta$ take the minimal value $L_0$. This qualitative mass is quasi-normalized since $L_1+L_1+L_4=L_{1+1+4}=L_6=L_{\max}$.

\noindent If we assume that the conditioning event is the proposition $A\cup B$, i.e. the absolute truth is in $A\cup B$, the hyper-power set decomposition (HPSD) is obtained as follows: $D_1$ is formed by all parts included in $A\cup B$, i.e. $D_1=\{A\cap B, A, B, A\cup B, B\cap D, A\cup (B\cap D), (A\cap B)\cup (B\cap D)\}$, $D_2$ is the set generated by $\{(C,D),\cup,\cap\} \setminus \emptyset=\{C,D,C\cup D, C\cap D\}$, and $D_3=\{A\cup C, A\cup D, B\cup C, B\cup D, A\cup B\cup C, A\cup (C\cap D), \ldots\}$.\\

\noindent The qualitative mass of element $D$ is transferred to $D\cap(A\cup B)=B\cap D$ according to the model, since $D$ is in the set $D_2\cap D_3$ and the largest element $X$ in $D_1$ which is contained by element $D$ is $B\cap D$.  Whence $qm_{QBCR1}(B\cap D | A\cup B) = L_4$, while $qm_{QBCR1}(D|A\cup B) = L_0$. The qualitative mass of element $C$, which is in $D_2\cup D_3$, but $C$ has no intersection with $A\cup B$ (i.e. the intersection is empty), is transferred to the whole $A\cup B$.  Whence $qm_{QBCR1}(A\cup B|A\cup B)=L_1$, while $qm_{QBCR1}(C|A\cup B)=L_0$. Since the truth is in $A\cup B$, then the qualitative masses of the elements $A$ and $B$, which are included in $A\cup B$, are not changed in this example, i.e. $qm_{QBCR1}(A| A\cup B)=L_1$ and $qm_{QBCR1}(B|A\cup B)=L_0$. One sees that the resulting qualitative conditional mass, $qm_{QBCR1}(.)$ is also quasi-normalized since $$L_4+L_0+L_1+L_0+L_1+L_0=L_6=L_{\max}$$
\noindent
In summary, one gets the following qualitative conditioned masses with QBCR1\footnote{Only non minimal linguistic values are given here since all the masses of other elements (i.e. non focal elements) take by default the value $L_0$.}:

\begin{align*}
qm_{QBCR1}(B\cap D|A\cup B)&=L_4\\
qm_{QBCR1}(A\cup B|A\cup B)& =L_1\\
qm_{QBCR1}(A| A\cup B)&=L_1
\end{align*}

Analogously to QBCR1, with QBCR2 the qualitative mass of the element $D$ is transferred to $D\cap (A\cup B)=B\cap D$ according to the model, since $D$ is in $D_2\cup D_3$ and the largest element $X$ in $D_1$ which is contained by $D$ is $B\cap D$.  Whence $qm_{QBCR2}(B\cap D|A\cup B) = L_4$, while $qm_{QBCR2}(D|A\cup B) = L_0$. But, differently from QBCR1, the qualitative mass of $C$, which is in $D_2\cup D_3$, but $C$ has no intersection with $A\cup B$ (i.e. the intersection is empty), is transferred $A$ only since $A \in A\cup B$ and $qm_1(A)$ is different from zero (while other sets included in $A\cup B$ have the qualitative mass equal to $L_0$).  Whence $qm_{QBCR2}(A|A\cup B)=L_1+L_1=L_2$, while $qm_{QBCR2}(C|A\cup B)=L_0$. Similarly, the resulting qualitative conditional mass, $qm_{QBCR2}(.)$ is also quasi-normalized since $L_4+L_0+L_2+L_0 =L_6=L_{\max}$. Therefore the result obtained with QBCR2 is:
\begin{align*}
qm_{QBCR2}(B\cap D|A\cup B)&=L_4\\
qm_{QBCR2}(A|A\cup B)&=L_2
\end{align*}

\subsection{Example 2}

Let's consider a more complex example related with military decision support. We assume that the frame $\Theta=\{A,B,C,D\}$ corresponds to the set of four regions under surveillance because these regions are known to potentially protect some dangerous enemies. The linguistic labels used for specifying qualitative masses belong to $L=\{L_0,L_1,L_2,L_3,L_4,L_5,L_6\}$. Let's consider the following prior qualitative mass $qm(.)$ defined by:
$$qm(A)=L_1, qm(C)=L_1, qm(D)=L_4$$
\noindent
All other masses take the value $L_0$. This qualitative mass is quasi-normalized since 
$L_1+L_1+L_4 = L_{1+1+4} = L_6 = L_{\max}$.

\noindent
We assume that the military headquarter has decided to bomb in priority region $D$ because there was a high qualitative belief on the presence of enemies in zone $D$ according to the prior qbba $qm(.)$. But let's suppose that after bombing and verification, it turns out that the enemies were not in $D$.
The important question the headquarter is now face to is on how to revise its prior qualitative belief $qm(.)$ knowing that the absolute truth is now not in $D$, i.e. $\bar{D}$ (the complement of $D$) is absolutely true.  The problem is a bit different from the previous one since the conditioning term $\bar{D}$ in this example does not belong to the hyper-power set $D^\Theta$. In such case, one has to work actually directly on the super-power set\footnote{The super-power $S^\Theta$ is the Boolean algebra $(\Theta,\cap,\cup,\mathcal{C})$ where $\mathcal{C}$ denotes the complement, while hyper-power set $D^\Theta$ corresponds to $(\Theta,\cap,\cup)$.} as proposed in \cite{DSmTBook_2006} (Chap. 8). $\bar{D}$ belongs to $D^\Theta$ only if Shafer's model (or for some other specific hybrid models -  see case 2 below) is adopted, i.e. when region $D$ has no overlap with regions $A$, $B$ or $C$. The truth is not in $D$ is in general (but with Shafer's model or with some specific hybrid models) not equivalent to the truth is in $A\cup B\cup C$ but with the truth is in $\bar{D}$. That's why the following two cases need to be analyzed:\\

\begin{itemize}
\item {\bf{Case 1}}: $ \bar{D}\neq A\cup B \cup C.$

If we consider the model represented in Figure \ref{venn2}, then it is clear that $\bar{D}\neq A\cup B \cup C$.
\begin{figure}[h]
\centering
{\tt \setlength{\unitlength}{1pt}
\begin{picture}(90,90)
\thinlines    
\put(20,60){\circle{40}} 
\put(40,60){\circle{40}} 
\put(30,40){\circle{40}} 
\put(50,50){\circle{40}} 
\put(-5,84){\vector(1,-1){10}}
\put(-13,84){$A$}
\put(64,84){\vector(-1,-1){10}}
\put(65,84){$B$}
\put(2,20){\vector(1,1){10}}
\put(-7,10){$C$}
\put(80,50){\vector(-1,0){10}}
\put(81,48){$D$}
\end{picture}}
\vspace{2mm}
\caption{Venn Diagram for case 1}
\label{venn2}
 \end{figure}
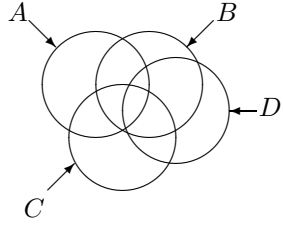

The Super-Power Set Decomposition (SPSD) is the following:
\begin{itemize}
\item if the truth is in $A$, then $D_1$ is formed by all non-empty parts of $A$;
\item $D_2$ is formed by all non-empty parts of $\bar{A}$;
\item $D_3$ is formed by what's left, i.e. $D_3 =(S^\Theta \setminus \{\emptyset\}) \setminus (D_1\cup D_2)$; thus $D_3$ is formed by all elements from $S^\Theta$ which have the form of unions
of some element(s) from $D_1$ and some element(s) from $D_2$, or by all
elements from $S^\Theta$ that overlap $A$ and $\bar{A}$.
\end{itemize}
In our particular example: $D_1$ is formed by all non-empty parts of $\bar{D}$; $D_2$ is formed by all non-empty parts of $D$; $D_3 =\{A, B, C, A\cup D, B\cup D, A\cup B, \ldots\}$.\\

\begin{itemize}
\item[a)]
Using QBCR1: one gets:
\begin{align*}
qm_{QBCR1}(A\cap \bar{D}|\bar{D})&=L_1\\
qm_{QBCR1}(C\cap \bar{D}|\bar{D})&=L_1\\
qm_{QBCR1}(\bar{D}|\bar{D})&=L_4
\end{align*}
\item[b)] Using QBCR2: one gets
\begin{align*}
qm_{QBCR2}(A\cap \bar{D}|\bar{D})&=L_1+\frac{1}{2}L_4 = L_1+ L_{[\frac{4}{2}]}=L_3\\
qm_{QBCR2}(C\cap \bar{D}|\bar{D})&=L_1+\frac{1}{2}L_4=L_3
\end{align*}
\end{itemize}

Note that with both conditioning rules, one gets quasi-normalized qualitative belief masses.
The results indicate that zones $A$ and $C$ have the same level of qualitative belief after the conditioning which is normal. QBRC1 however, which is more prudent, just commits the higher belief to the whole zone $A\cup B\cup C$ which represents actually the less specific information, while QBRC2 commits equal beliefs to the restricted zones $A\cap \bar{D}$ and $C\cap \bar{D}$ only. As far as only the minimal surface of the zone to bomb is concerned (and if zones $A\cap \bar{D}$ and $C\cap \bar{D}$ have the same surface), then a random decision has to be taken between both possibilities. Of course some other military constraints need to be taking into account in the decision process in such situation if the random decision choice is not preferred.\\

\item {\bf{Case 2}}: $ \bar{D}= A\cup B \cup C.$
This case occurs only when  $D\cap (A\cup B\cup C)=\emptyset$ as for example to the following model\footnote{This condition is obviously also satisfied for Shafer's model, i.e. when all regions are well separate/distinct.}. In this second case, "the truth is not in $D$" is equivalent to "the truth is in $A\cup B \cup C$". The decomposition is the following:
$D_1$ is formed by all non-empty parts of $A\cup B\cup C$; $D_2 = \{D\}$; $D_3 = \{A\cup D, B\cup D, C\cup D, A\cup B\cup D, A\cup C\cup D, B\cup C\cup D,
A\cup B\cup C\cup D, (A\cap B)\cup D, (A\cap B\cap C)\cup D, ...\}$.

\begin{figure}[h]
\centering
{\tt \setlength{\unitlength}{1pt}
\begin{picture}(90,90)
\thinlines    
\put(20,60){\circle{40}} 
\put(40,60){\circle{40}} 
\put(30,40){\circle{40}} 
\put(90,50){\circle{40}} 
\put(-5,84){\vector(1,-1){10}}
\put(-13,84){$A$}
\put(64,84){\vector(-1,-1){10}}
\put(65,84){$B$}
\put(2,20){\vector(1,1){10}}
\put(-7,10){$C$}
\put(120,50){\vector(-1,0){10}}
\put(121,48){$D$}
\end{picture}}
\vspace{2mm}
\caption{Venn Diagram for case 2 }
\label{venn3}
 \end{figure}
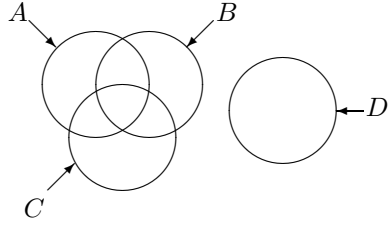

\begin{itemize}
\item[a)]
Using QBCR1: one gets
\begin{align*}
qm_{QBCR1}(A|\bar{D})&=L_1\\
qm_{QBCR1}(C|\bar{D})&=L_1\\
qm_{QBCR1}(A\cup B \cup C|\bar{D})&=L_4
\end{align*}
\item[b)]
Using QBCR2: one gets
\begin{align*}
qm_{QBCR2}(A|\bar{D})&=L_3\\
qm_{QBCR2}(C|\bar{D})&=L_3\\
\end{align*}
\end{itemize}

\end{itemize}

Same concluding remarks as for case 1 can be drawn for the case 2. Note that in this case, there is uncertainty in the decision to bomb zone $A$ or zone $C$ because they have the same supporting belief. The only difference with respect to case 1, it that the zone to be bomb (whatever the one chosen - $A$ or $C$) will remain larger than in case 1 because $D$ has no intersection with $A$, $B$ and $C$ for this model.

\subsection{Example 3}

Let's modify the previous example for examining what happens when using  an {\it{unconventional}} bombing strategy. Here we still consider four zones under surveillance, i.e. $\Theta=\{A,B,C,D\}$ and $L=\{L_0,L_1,L_2,L_3,L_4,L_5,L_6\}$ but with the following prior quasi-normalized qualitative basic belief mass $qm(.)$:
$$qm(A)=L_1, qm(C)=L_3, qm(D)=L_2$$

\noindent All other qualitative masses take the value $L_0$. Such prior suggests normally/rationally to bomb in priority the zone $C$ since it is the one carrying the higher belief on the location of enemies. But for some unknown reasons (military, political or whatever) let's assume that the headquarter has finally decided to bomb $D$ first. Let's examine how will be revised the prior $qm(.)$ with QBCR1 and QBCR2 in such situation for the two cases:\\

\begin{itemize}
\item {\bf{Case 1}}: $ \bar{D}\neq A\cup B \cup C.$
\begin{itemize}
\item[a)]
Using QBCR1: $qm(A)=L_1$ is transferred to $A\cap  \bar{D}$, since $A\cap  \bar{D}$ is the largest element from $ \bar{D}$ which is included in $A$, so we get $qm_{QBCR1}(A\cap  \bar{D}|  \bar{D})=L_1$;
and similarly $qm(C)=L_3$ is transferred to $C\cap  \bar{D}$, since $C\cap  \bar{D}$ is the largest element from $\bar{D}$ which is included in $C$, so we get $qm_{QBCR1}(C\cap  \bar{D}|  \bar{D})=L_3$;
Also, $qm_2(D)=L_2$ is transferred to $ \bar{D}$ since no element from $\bar{D}$ is included in $D$, therefore $qm_{QBCR1}( \bar{D}| \bar{D})=L_2$. Analogously, this qualitative conditioned mass $qm_{QBCR1}(.)$ is quasi-normalized since $L_1+L_3+L_2=L_6=L_{\max}$. In summary, with QBCR1 one gets in this case:
\begin{align*}
qm_{QBCR1}(A\cap \bar{D}|\bar{D})&=L_1\\
qm_{QBCR1}(C\cap \bar{D}|\bar{D})&=L_3\\
qm_{QBCR1}(\bar{D}|\bar{D})&=L_2
\end{align*}

\noindent
\item[a)] Using QBCR2: $qm(A)=L_1$ is transferred to $A\cap  \bar{D}$, and $qm(C)=L_3$ is transferred to $C\cap  \bar{D}$. Since no qualitative focal element exists in $\bar{D}$, then $qm(D)=L_2$ is transferred to $ \bar{D}$, and we get the same result as for QBCR1.\\
\end{itemize}

\item {\bf{Case 2}}: $ \bar{D}= A\cup B \cup C.$

\begin{itemize}
\item[a)] Using QBCR1: the qualitative masses of $A$, $B$, $C$ do not change since they are included in $A\cup B\cup C$ where the truth is. The qualitative mass of $D$ becomes {\it{zero}} (i.e. it takes the linguistic value $L_0$) since $D$ is outside the truth, and $qm(D)=L_2$ is transferred to $A\cup B\cup C$. Hence:
\begin{align*}
qm_{QBCR1}(A|\bar{D})&=L_1\\
qm_{QBCR1}(C|\bar{D})&=L_3\\
qm_{QBCR1}(A\cup B \cup C|\bar{D})&=L_2
\end{align*}
\noindent This resulting qualitative conditional mass is also quasi-normalized.\\
\item[b)] QBCR2, the qualitative mass of $D$ becomes (linguistically) {\it{zero}} since $D$ is outside the truth, but now $qm(D)=L_2$ is equally split to $A$ and $C$ since they are the only qualitative focal elements from $D_1$ which means all parts of $A\cup B\cup C$, therefore each of them $A$ and $C$ receive $(1/2)L_2=L_1$. Hence: 
$$
qm_{QBCR2}(A|\bar{D})=L_1 + (1/2)L_2  = L_1 + L_{2/2} = L_1 + L_1 = L_2
$$
$$
qm_{QBCR1}(C|\bar{D})=L_3 + (1/2)L_2 = L_3 + L_{2/2} = L_3 + L_1 = L_4
$$
\noindent
Again, the resulting qualitative conditional mass is quasi-normalized.\\
\end{itemize}

As concluding remark, we see that even if a {\it{unconventional}} bombing strategy is chosen first, the results obtained by QBCR rules 1 or 2 are legitimate and coherent with intuition since they commit the higher belief in either $C\cap \bar{D}$ (case 1) or $C$ (case 2) which is normal because the prior belief mass in $C$ was the higher one before bombing $D$.

\end{itemize}

\subsection{Example 4}

Let's complicate a bit the previous example by working directly with a prior $qm(.)$ defined on the super-power set $S^\Theta$ (see the previous Footnote 3), i.e. the complement is allowed among the set of propositions to deal with. As previously, we consider four zones under surveillance, i.e. $\Theta=\{A,B,C,D\}$ and $L=\{L_0,L_1,L_2,L_3,L_4,L_5,L_6\}$. The following prior qualitative basic belief mass
$qm(.)$ is extended from the hyper-power set to the super-power set, i.e. $qm(.): S^\Theta \rightarrow L$:
$$qm(A)=L_1, \qquad qm(C)=L_1,  \qquad qm(D)=L_2$$
$$qm(C\cup D)=L_1,  \qquad qm(C\cap \bar{D})=L_1$$

\noindent
All other qualitative masses take the value $L_0$. This qualitative mass is quasi-normalized since 
$$L_1+L_1+L_2+L_1+L_1 = L_{1+1+2+1+1} = L_6 = L_{\max}$$

\noindent
We assume that the military headquarter has decided to bomb in priority region $D$ because there was a high qualitative belief on the presence of enemies in $D$ according to the prior qbba $qm(.)$. But after bombing and verification, it turns out that the enemies were not in $D$ (same scenario as for example 2).
Let's examine the results of the conditioning by the rules QBCR1 and QBCR2 for the cases 1 and 2:\\

\begin{itemize}
\item {\bf{Case 1}}: $ \bar{D}\neq A\cup B \cup C.$

\begin{itemize}
\item[a)]
Using QBCR1: $qm(A)=L_1$ is transferred to $A\cap  \bar{D}$, since $A\cap  \bar{D}$ is the largest element (with respect to inclusion) from $ \bar{D}$ which is included in $A$. $qm(C)=L_1$ is similarly transferred to $C\cap  \bar{D}$, since $C\cap  \bar{D}$ is the largest element from $\bar{D}$ which is included in $C$. $qm(C\cup D)=L_1$ is also transferred to $C\cap \bar{D}$ since $C\cap \bar{D}$ is the largest element from $\bar{D}$ which is included in $C\cup D$. $qm(D)=L_2$ is transferred to $\bar{D}$ since no element from $\bar{D}$  is included in $D$. In summary, we get:
\begin{align*}
qm_{QBCR1}(A\cap \bar{D}|\bar{D})&=L_1\\
qm_{QBCR1}(C\cap \bar{D}|\bar{D})&=qm(C\cap \bar{D})+qm(C) +qm(C\cup D) = L_1+L_1+L_1=L_3\\
qm_{QBCR1}(\bar{D}|\bar{D})&=L_2
\end{align*}
\noindent
All others are equal to $L_0$. The resulting qualitative conditioned mass is quasi-normalized since $L_1+L_3+L_2=L_6=L_{\max}$.

\item[b)]
\noindent
Using  QBCR2: Similarly as for QBCR1, $qm(A)=L_1$ is transferred to $A\cap  \bar{D}$; also $qm(C)=L_1$ and $qm(C\cup D)=L_1$ are transferred to $C\cap  \bar{D}$. But now, differently, $qm(D)=L_2$ is equally split to the focal elements of $\bar{D}$, but only $C\cap \bar{D}$ is focal for $\bar{D}$, so $C\cap \bar{D}$ receives the whole qualitative mass of $D$. Finally we get:
\begin{align*}
qm_{QBCR2}(A\cap \bar{D}|\bar{D})&=L_1\\
qm_{QBCR2}(C\cap \bar{D}|\bar{D})&=qm(C\cap \bar{D})+qm(C) +qm(C\cup D)+qm(D) = L_1+L_1+L_1+L_2=L_5\\
\end{align*}
\noindent
All others are equal to $L_0$. The resulting qualitative conditioned mass is quasi-normalized since $L_1+L_5=L_6=L_{\max}$.\\
\end{itemize}

The results obtained by QBCR1 and QBCR2 are coherent with rational human reasoning since after bombing zone $D$ we get, in such case, a higher belief in finding enemies in $C\cap \bar{D}$ than in $A\cap \bar{D}$ which is normal due to the prior information we had before bombing $D$. QBRC2 is more specific than QBRC1. Say differently, QBRC1 is more prudent than QBRC2 in the revision of the masses of belief.\\

\item {\bf{Case 2}}: $ \bar{D}= A\cup B \cup C.$

\begin{itemize}
\item[a)]
Using QBCR1: $qm(C\cup D)=L_1$ is transferred to $C$ since $C$ is the largest element (with respect to inclusion) from $A\cup B\cup C$ which is included in $C\cup D$. $qm(C\cap \bar{D})=qm(C)$ since $C\cap (A\cup B\cup C)=C$. $qm(D)=L_2$ is transferred to $A\cup B\cup C$ since no element from $A\cup B\cup C$ is included in $D$, so the qualitative mass of $D$ becomes {\it{zero}} (i.e. it takes the linguistic value $L_0$). Thus we finally obtain:
\begin{align*}
qm_{QBCR1}(A|\bar{D})&=L_1\\
qm_{QBCR1}(C|\bar{D})&=qm(C)+qm(C\cup D) + qm(C\cap\bar{D}) = L_1+L_1+L_1=L_3\\
qm_{QBCR1}(A\cup B \cup C|\bar{D})&=L_2
\end{align*}
\noindent
All others are equal to $L_0$. The resulting qualitative conditioned mass is quasi-normalized since $L_1+L_3+L_2=L_6=L_{\max}$.

\item[b)]
Using QBCR2: $qm(C\cup D)=L_1$ and $qm(C\cap \bar{D})=L_1$ are similarly as in QBRC1 transferred to $C$. But $qm(D)=L_2$ is equally split among the focal qualitative elements of $ \bar{D}= A\cup B \cup C$, which are $A$ and $C$, so each of them receive $1/2\cdot L_2=L_{2/2}=L_1$. Whence
$$
qm_{QBCR2}(A|\bar{D})=qm(A)+\frac{1}{2}qm(D) = L_1+ \frac{1}{2} L_2=L_1+L_1=L_2 
$$
$$
qm_{QBCR1}(C|\bar{D})=[qm(C)+qm(C\cup D)  + qm(C\cap \bar{D})] + \frac{1}{2}qm(D)
 = [L_1+L_1+L_1] + L_1=L_4$$
\noindent
All others are equal to $L_0$. The resulting qualitative conditioned mass is quasi-normalized since $L_2+L_4=L_6=L_{\max}$.\\
\end{itemize}
The results obtained by QBCR1 and QBCR2 are again coherent with rational human reasoning since after bombing zone $D$ we get, in such case, a higher belief in finding enemies in $C$ than in $A$ which is normal due to the prior information we had before bombing $D$ and taking into account the constraint of the model.\\

\end{itemize}

\section{Conclusions}

In this paper, we have designed two Qualitative Belief Conditioning Rules in order to revise qualitative basic belief
assignments and we presented some examples to show how they work. QBCR1 is more prudent than QBCR2 because the revision of the belief is done in a less specific transfer than for QBCR2.  We use it when we are less confident in the source. While QBCR2 is more optimistic and refined; we use it when we are more confident in the source. Of course, the qualitative conditioning process is less precise than its quantitative counterparts because it is based on a rough approximation, as it normally happens when working with linguistic labels. Such qualitative methods present however some interests for manipulating information and beliefs expressed in natural language by human experts and can be helpful for high-level decision support systems.

\end{document}